\pdfoutput=1

\documentclass[11pt]{article}

\usepackage{EMNLP2023}

\usepackage{times}
\usepackage{latexsym}

\usepackage[T1]{fontenc}

\usepackage[utf8]{inputenc}

\usepackage{microtype}

\usepackage{inconsolata}

\usepackage{url}
\usepackage{xcolor}
\usepackage{booktabs}
\usepackage{graphicx}
\usepackage{hyperref}
\usepackage{todonotes}
\usepackage{multirow}
\usepackage{svg}

\newcommand{\up}{\textcolor{green}{$\uparrow$}}
\newcommand{\down}{\textcolor{red}{$\downarrow$}}

\definecolor{c_neighborhood}{RGB}{218, 234, 212}
\definecolor{c_head}{RGB}{207, 226, 243}
\definecolor{c_tail}{RGB}{243, 159, 166}

\title{Better Together: Enhancing Generative Knowledge Graph Completion with Language Models and Neighborhood Information}

\author{
Alla Chepurova$^1$\quad
Aydar Bulatov$^1$\quad
Yuri Kuratov$^{1,2}$\quad
Mikhail Burtsev$^3$\\\\
$^1$Neural Networks and Deep Learning Lab, MIPT, Dolgoprudny, Russia \\
$^2$AIRI, Moscow, Russia\\
$^3$London Institute for Mathematical Sciences, London, UK\\
\texttt{chepurova@deeppavlov.ai}, \texttt{\{bulatov.as,yurii.kuratov\}@phystech.edu}, \texttt{mb@lims.ac.uk}\\\\
}

\begin{document}
\maketitle
\begin{abstract}
Real-world Knowledge Graphs (KGs) often suffer from incompleteness, which limits their potential performance.
Knowledge Graph Completion (KGC) techniques aim to address this issue.
However, traditional KGC methods are computationally intensive and impractical for large-scale KGs, necessitating the learning of dense node embeddings and computing pairwise distances.
Generative transformer-based language models (e.g., T5 and recent KGT5) offer a promising solution as they can predict the tail nodes directly.
In this study, we propose to include node neighborhoods as additional information to improve KGC methods based on language models.
We examine the effects of this imputation and show that, on both inductive and transductive Wikidata subsets, our method outperforms KGT5 and conventional KGC approaches. We also provide an extensive analysis of the impact of neighborhood on model prediction and show its importance. Furthermore, we point the way to significantly improve KGC through more effective neighborhood selection.
\end{abstract}

\section{Introduction}

Knowledge graphs (KG) represent structured information in the form of a directed multi-relational graph with entities as the graph's nodes and relations between entities as the graph's edges. Entities in such graphs could be both real-world objects or abstract concepts. KGs are represented as a set of triplets with representation entities and a relation between them, i.e. (head entity, relation, tail entity) or (h, r, t) for short. Due to KGs' recent rapid development, they are now widely used in a variety of applications, including data mining, information retrieval, question answering, and recommendation systems. However, most real-world knowledge graphs suffer from incompleteness. Knowledge graph completion (KGC) algorithms attempt to fill in the gaps in the KG in order to solve this problem.


\begin{figure}[h]
  \centering
    \includegraphics[width=\linewidth]{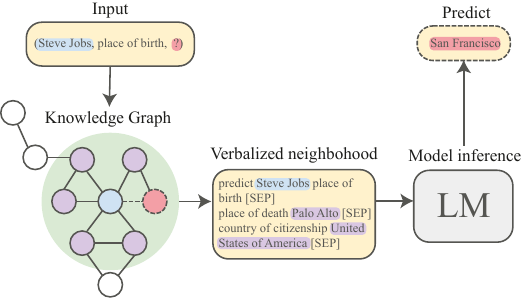}
  \caption{\textbf{Graph completion pipeline.} We extract and verbalize \colorbox{c_neighborhood}{neighborhood} of a \colorbox{c_head}{triplet head node}  from the Knowledge Graph (KG). Language Model uses verbalized input triplet and neighborhood to make \colorbox{c_tail}{predictions}. Predictions can be used for question answering or KG completion.}
  \label{fig:overall_pipeline}
  \vskip -0.2in
\end{figure}

KG link prediction commonly involves learning knowledge graph embeddings (KGE), which capture structural and semantic information about KG nodes, relations, and their neighborhoods in a way that a simple distance-based function would be able to distinguish real triplets from false ones.
However, building a unique embedding for each node and relation leads to a linear increase in model size and inference time as the KG size grows. Another drawback is that KG updating requires full re-training of KGEs, rather than simply learning new embeddings for updated entities or relations.

Application of language models (LMs) to KGC tasks is a promising direction~\citep{pan2023unifying}. State-of-the-art performance in most natural language processing (NLP) tasks has recently been demonstrated by pre-trained language models like BERT~\citep{devlin-etal-2019-bert}
and more recent large or foundational language models~\citep{gpt3, chowdhery2022palm, touvron2023llama}.
Language models-based approaches, such as KG-BERT~\citep{yao2019kgbert}, BERTRL~\citep{zha2022bertrl}, KGT5~\citep{kgt5} and GenKGC~\citep{xie2022discrimination} consider a triplet in the KG as a text sequence.
While distance-based methods incorporate more structural information in dense embeddings, language models tend to capture more semantic information contained in both relations and entities of KG, enabling such models to embed previously unseen entities and operate on new facts based on semantic information contained in triplets textual representation.
Thus, KGT5 positions KG link prediction as sequence-to-sequence tasks. KGT5 model trained to generate target node using textually verbalized triplets achieves competitive performance and offers a scalable and simple framework for KGC and QA tasks.

Current methods based on language models (KG-BERT, BERTRL, KGT5, and GenKGC) rely only on information that can be extracted from LMs parameters. However, since KG itself is available, we build a method that relies on both internal language model knowledge and external information from KG. For the link prediction task ($h$, $r$, $?$), we propose to extract nodes and relations adjacent to $h$ from the KG, i.e. the neighborhood. Node $h$ with its neighborhood \textit{better together} represents $h$ itself: neighborhood may contain clues such as Palo Alto and United States of America for the triplet (Steve Jobs, place of birth, ?) on Figure~\ref{fig:overall_pipeline}.

Though some efforts have been made to incorporate neighborhood into the KGC algorithms ~\cite{chen2020hitter}, our proposed pipeline inherits the benefits of generative LLMs compared to KGE approaches: (1) scalability and size of the model, (2) applicability to both transductive and inductive KGs due to an ability to generalize unseen entities, (3) no need to rank all possible candidate triplets due to direct generation of tail entity. 
Also, a concurrent work by the KGT5 authors, KGT5-context~\cite{kochsiek2023friendly}, proposed a similar idea of integrating node neighborhood in the context of the generative LM model, supporting the main concerns and results of our study.

 \textbf{Our contributions are as follows:}
 (1) we propose a generative graph completion method which effectively integrates language models with graph neighborhood information.
 (2) We show that incorporating graph information into multiple generative language models achieve strong performance improvement of link prediction in both transductive and inductive settings. Moreover, our approach achieves state-of-the-art results on the inductive ILPC dataset.
 (3) We analyze and evaluate the impact of adding neighborhood information on the performance of the models.
 
\section{Methods}

One can represent KG more formally as $\mathcal{G} = (\mathcal{T}, \mathcal{E}, \mathcal{R})$, where $\mathcal{E}$ is a set of entities, $\mathcal{R}$ is a set of relations and $\mathcal{T}$ is a set of triplets $\mathcal{T} \in \mathcal{E} \times \mathcal{R} \times \mathcal{E}$.
Therefore, the formulation of the KGC task is the following: for each triplet $\mathcal{T}_{i} = (h, r, t)$, where $h \in \mathcal{E}, t \in \mathcal{E}, r \in \mathcal{R}$, answer both $(h, r, ?)$ and $(?, r, t)$ queries on the graph $\mathcal{G} /\mathcal{T}_{i}$. The KGC tasks can have different variations depending on the KG type: inductive and transductive. In a transductive setting the KGC task is formulated as follows: given a train KG $\mathcal{G}_{train} = (\mathcal{T}_{train}, \mathcal{E}, \mathcal{R})$ train the model to predict on the inference KG $\mathcal{G}_{inference} = (\mathcal{T}_{inference}, \mathcal{E}, \mathcal{R})$, therefore in the train and inference KG there is a same set of entities and relations. In inductive KGC the model is trained on the KG $\mathcal{G}_{train} = (\mathcal{T}_{train}, \mathcal{R}, \mathcal{E}_{train}$) and predicts the inference KG $\mathcal{G}_{inference} = (\mathcal{T}_{inference}, \mathcal{R}, \mathcal{E}_{inference})$, so the trained model observed only on the entities from train subgraph of KG.

We followed the KGT5~\citep{kgt5} approach to the KGC task using an encoder-decoder Transformer based on T5~\citep{raffel2020t5}. The link prediction task was posed as follows: for each triplet $(entity_1, relation, entity_2)$ from KG we had to predict the tail in the queries $(entity_1, relation, ?)$ and $(entity_2, relation^{-1}, ?)$.
To generate input for the model, we converted such queries into a text form. Apart from the query verbalization only as in KGT5, we formed a neighborhood for the node and relations for each triplet in KG. Thus, verbalized queries and their neighborhoods in the form of text were fed to the input of the model for training (see Figure~\ref{fig:overall_pipeline}) and later to predict incomplete triplets. We form a neighborhood of nodes that have an edge with the triplet head node (1-hop neighbors). The number of adjacent nodes may be too large to fit into the context size of selected language models. Therefore, we sort them based on relation semantic similarity. We take only those verbalized nodes and relations that fit into the max sequence length. We provide more details on neighborhood formation in Appendix~\ref{sec:neighborhood_formation} and verbalization in Appendix~\ref{sec:triplets_verbalization}.

To evaluate our approach to link prediction tasks we used publically available transductive and inductive KGs. Specifically, we used the Wikidata5M~\citep{wang2021kepler} dataset, which is one of the largest publicly available transductive KG up-to-date and large and small versions of ILPC dataset~\citep{ilpc}, the largest fully inductive benchmark. ILPC inference graph size is challenging for modern GNN methods with inductive capabilities ~\cite{ilpc}. Apart from being challenging to solve by common approaches, the ILPC KGs are much more sparse than the transductive Wikidata5M. The graph size and sparseness of the data make the task hard for existing methods and result in modest performance metrics. Still, there were no existing solutions that beat the baseline provided in the ILPC paper. Both used datasets are the subsets of the real-world KG -- Wikidata. Datasets' statistics are provided in the Table~\ref{table:dataset_statistics}).

We compared models with and without neighborhoods to evaluate our hypothesis that neighborhoods help to \textit{better together} represent graph node. In addition, unlike the KGT5, we experimented with different model configurations, both fine-tuning an existing pre-trained model and training from scratch as common knowledge obtained during language models pre-training might be useful on real-world knowledge graphs like Wikidata.

We used the T5-small with 60M parameters as our base model with both pretrained and randomly initialized weights to evaluate the effect of pretraining of the language model on the KG task.
We added additional "[SEP]" tokens to the tokenizer with 30K tokens vocabulary to separate query and neighborhood. The model was trained using AdamW optimizer~\citep{loshchilov2018adamw}, 1e-5 learning rate, 10\% dropout, and batch size equals 320. The number of training steps was equal to 5M for the Wikidata5M dataset and 2M for all others. For multi-gpu training, we used the Horovod parallelization framework~\citep{horovod}. In cases where a batch size equal to 320 did not fit GPU, we used gradient accumulation steps. We also benchmarked OpenAI \texttt{gpt-3.5-turbo}\footnote{\texttt{gpt-3.5-turbo-0613}: \url{https://platform.openai.com/docs/models/gpt-3-5}} zero-shot link
prediction and evaluate the benefits of a neighborhood on LLM (see Appendix~\ref{sec:gpt}). The code for all the described experiments is available at the link\footnote{\url{https://github.com/screemix/kgc-t5-with-neighbors}}.

As KGT5, our model produces a probability distribution over all possible tokens in the vocabulary at each step of decoding. Such distribution is penalized for not matching the actual distribution, i.e., for the difference between the real target token at position $i$ and the generated token at step $i$ using CE loss.
Such an approach is principally different from distance-based methods as it does not require generating negative samples to make them further from the target prediction for each training query. In this way, training iteration becomes independent of the KG size.

To form predictions, we first verbalize the query  $(head, relation, ?)$ and pass it as an input to the generative model to produce an output sequence, which will be a verbalization of the target entity id.
For inference, we employed two strategies.
The first one was sampling $ k $ most probable sequences from the decoder.
We follow the KGT5~\citep{kgt5} setting and use a sample size of 50. We apply this approach to Wikidata5M datasets as it is transductive. The second option for inference was to search for the closest entities in KG to the generated one. We apply the second approach to ILPC inductive datasets as language models have to generate previously unseen entities. More details on inference procedure are in Appendix~\ref{sec:inference}.

\begin{table*}[htp]
\caption{Adding a neighborhood persistently improves results on the link prediction task. We report Hits@k metrics on Wikidata5M, ILPC-small, and ILPC-large test sets. Arrows \up\down~ show the effect of the neighborhood on language models. Neighbors allow to achieve the best performance considering the number of model parameters.}
\vskip -0.1in
\label{tab:results}
\begin{center}
\begin{small}
\begin{sc}
\resizebox{\textwidth}{!}{
\begin{tabular}{@{}llllllllllll@{}}
\toprule
\multirow{2}{*}{Model} & \multicolumn{3}{c}{Wikidata5M} & \multicolumn{3}{c}{ILPC-small} & \multicolumn{3}{c}{ILPC-large}  &  \\
\cmidrule(lr){2-4}\cmidrule(lr){5-7}\cmidrule(lr){8-10}
& H@1 & H@3 & H@10 & H@1 & H@3 & H@10 & H@1 & H@3 & H@10 & \#params \\
\midrule
KGT5~\citep{kgt5} & 0.267 & 0.318 & 0.365 & - & - & - & - & - & - & 60M\\
TransE~\citep{transe} & 0.170 & 0.311 & 0.392 & - & - & - & - & - & - & 2,400M \\
ComplEx~\citep{complex} & 0.255 & - & 0.398 & - & - & - & - & - & - & 614M \\
KEPLER~\citep{wang2021kepler} & 0.173 & 0.224 & 0.277 & - & - & - & - & - & - & 125M \\
SimKGC~\citep{wang2022simkgc} & 0.313 & 0.376 & 0.441 & - & - & - & - & - & - & 220M \\
IndNodePiece~\citep{ilpc} & - & - & - & 0.007 & 0.022 & 0.092 & 0.037 & 0.054 & 0.125 & 15.5K \\
IndNodePieceGNN~\citep{ilpc} & - & - & - & 0.076 & 0.140 & 0.251 & 0.032 & 0.073 & 0.146 & 24K\\

\midrule
T5-small not pre-trained & 0.240 & 0.286 & 0.323 & - & - & - & - & - & - & 60M\\
T5-small not pre-trained + neighbors & 0.327\up & 0.363\up & 0.386\up & - & - & - & - & - & - & 60M\\
T5-small pre-trained & 0.205 & 0.248 & 0.282 & 0.073 & 0.117 & 0.177 & 0.112 & 0.137 & 0.194 & 60M\\
T5-small pre-trained + neighbors & 0.306\up & 0.342\up & 0.361\up & 0.076\up & 0.137\up & 0.209\up & 0.114\up & 0.139\up & 0.200\up & 60M\\
T5-base pre-trained & - & - & - & 0.073 & 0.130 & 0.199 & 0.110 & 0.136	& 0.198 & 220M\\
T5-base pre-trained + neighbors & - & - & - & 0.082\up & 0.142\up & 0.216\up & 0.116\up & 0.141\up & 0.215\up & 220M \\

\bottomrule
\end{tabular}
}
\end{sc}
\end{small}
\end{center}
\vskip -0.1in
\end{table*}

\section{Results}

The hypothesis behind our approach was that neighborhood information incorporated in a context fed to the language model should enhance link prediction of generative LM. Tables~\ref{tab:results} and \ref{tab:chatgpt} demonstrate the performance of our technique on Wikidata5M and ILPC datasets under various conditions, including training a T5 model with and without neighbors, using a pre-trained model or training a model from scratch, and zero-shot with \texttt{gpt-3.5-turbo}. 

These results support the hypothesis that the quality of link prediction increases with the use of neighbors in the context of the generative model in both transductive and inductive settings. Moreover, our approach demonstrated SOTA performance on the ILPC dataset, while on Wikidata we achieved the best result among generative link prediction models, which is still comparable to scores of SOTA graph and LM methods with many more parameters. Relatively small absolute values of scores on the ILPC dataset can be explained by its fully inductive setup and larger sparseness compared to Wikidata5M. Still, there are no publicly available solutions that exceed the baseline score provided by ILPC authors. Our solution overperforms the existing baseline, is elegantly simple, and more interpretable than GNN-based solutions. Experiments on pre-trained models revealed that pre-training speeds up convergence but degrades the quality of the model by a few points compared to the one trained from scratch. 

Also, we demonstrated that neighbors' information enhances performance across different generative models. 
This includes open-source models such as T5, as well as proprietary models like OpenAI GPTs that are distributed as a model-as-a-service.
Although both types of models have their own advantages and limitations, it should be noted that OpenAI GPTs are black box models that cannot be modified or interpreted further.
Usage of OpenAI GPTs is restricted in the case of more specialized or proprietary KGs that can store secure or non-public information.
In the context of our study, OpenAI's \texttt{gpt-3.5-turbo} has a much larger size and cost, compared to smaller and transparent T5, still our approach improves the performance of both models.

\begin{table}
\caption{Neighbor information improves zero-shot link prediction with OpenAI \texttt{gpt-3.5-turbo-0613}. *ILPC-Large results were obtained on 5000 random samples from the test set. }
\vskip -0.1in
\label{tab:chatgpt}
\begin{center}
\begin{small}
\begin{sc}
\resizebox{0.48\textwidth}{!}{
\begin{tabular}{@{}lllllll@{}}
\toprule
\multirow{2}{*}{Dataset} & \multicolumn{2}{c}{GPT} & \multicolumn{2}{c}{GPT+NEIGH}  \\
\cmidrule(lr){2-3}\cmidrule(lr){4-5}
& H@1 & H@3 & H@1 & H@3 \\
\midrule
Wikidata5M & 0.098 & 0.131 & 0.113\up & 0.157\up \\
ILPC-small & 0.079 & 0.107 & 0.084\up & 0.118\up  \\
ILPC-large* & 0.103 & 0.134 & 0.114\up & 0.150\up  \\
\end{tabular}
}
\end{sc}
\end{small}
\end{center}
\vskip -0.2in
\end{table}

To get an understanding of how proposed models learn to use neighboring triplets we compared predictions of models trained with and without graph information in the input. 
One possible case when neighbors help in link prediction is when a target entity is somehow mentioned in the model input. Because of neighborhood filtering, the target triplet never appears in the model input, however in some cases target entity can occur with different relations or as a substring of another entity in a neighborhood or even in the task itself (Table~\ref{tab:target_inclusion}). 
Despite the tail appearing in input only in a small portion of the test set, the neighborhood model learns to make use of such hints, outperforming the baseline T5 in all selected cases. The high solution rate of more than 4 out of 5 samples with hinted targets supports the importance of properly selected neighborhoods.

\begin{figure}[h]
  \centering
    \includegraphics[width=\linewidth]{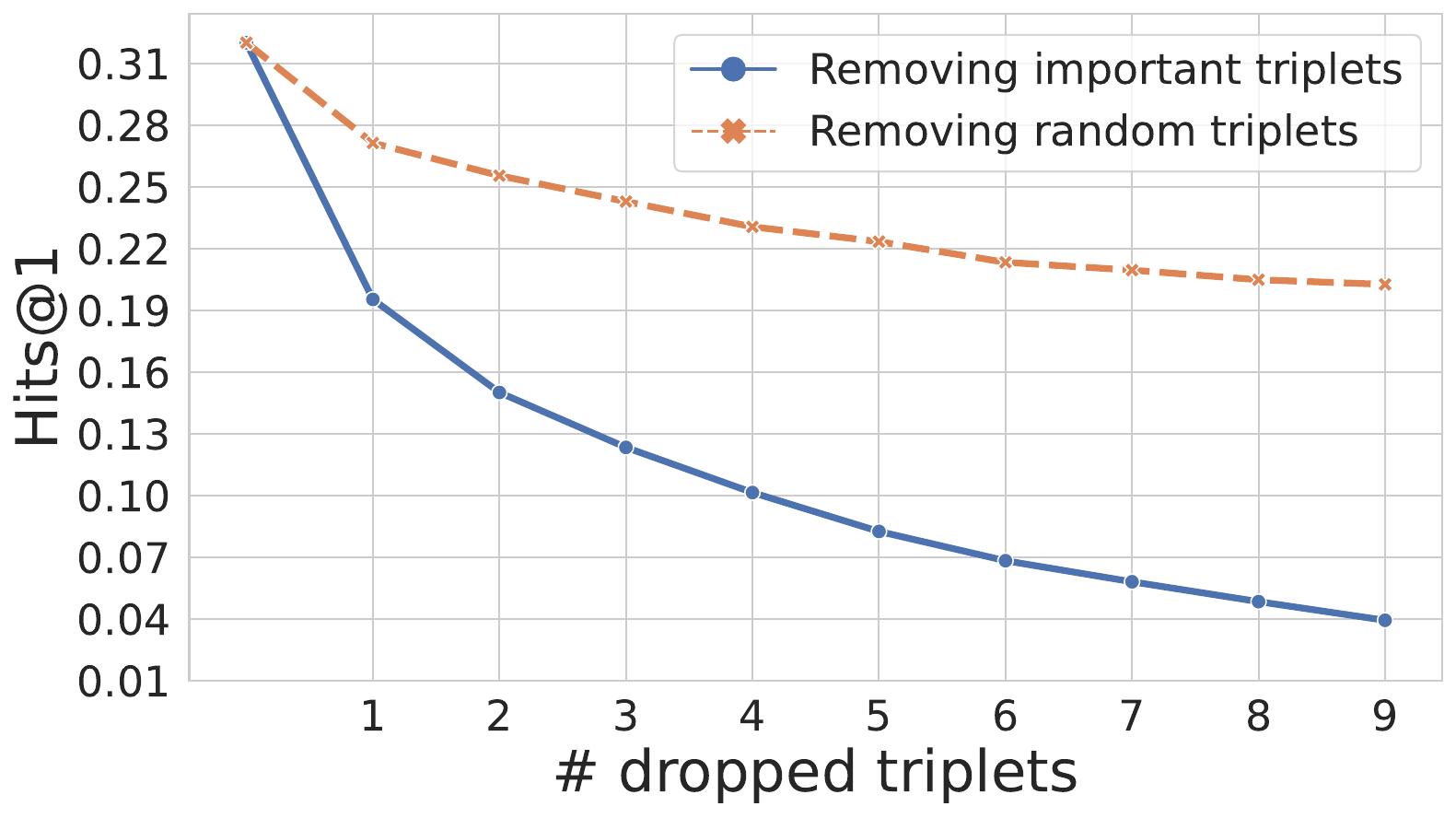}
  \caption{\textbf{KG neighborhood is critical for the model performance.} The plot shows how the quality of link prediction depends on the number of neighborhood triplets removed from the input context Wikidata5M T5-small not pre-trained model. The blue solid line corresponds to removing the most relevant triplets and the dashed red line represents random selection.}
  \label{fig:interptretation}
  \vskip -0.1in
\end{figure}

To evaluate the impact of neighborhoods, we analyzed triplets' importance by sorting them based on how they affect the probability of the target node prediction. 
As Figure~\ref{fig:interptretation} shows, a performance of the model quickly deteriorates with the removal of important triplets from the context. 
This implies potential avenues for improvement by prioritizing significant triplets selection prior to model training.
Moreover, we analyzed attention maps of the model train on Wikidata5M and found that it indeed pays attention to relevant neighbors, e.g. Appendix~\ref{sec:attention_maps}.

The analysis shows that adding relevant graph context information greatly enhances link prediction. Hence one way of improving performance is increasing neighborhood size by adding k-hop neighbors. Due to the quadratic complexity of attention and the limited input size of pretrained transformers, processing exponentially growing neighborhoods would be computationally expensive. To deal with large input sizes one can use efficient T5 architectures with input sizes up to 32k \citep{guo-etal-2022-longt5} and 64k \citep{ainslie2023colt5} tokens or recurrent approaches \citep{rmt_2022,hutchins2022blockrecurrent} that can sequentially process effectively unlimited neighborhood sizes. Another potential way of subgraph selection is to add relevant but more distant knowledge graph information, such as random walks or more sophisticated methods \citep{das2018minerva}.

\begin{table}
  \caption{Adding relevant neighbors improves link prediction for T5-small on the Wikidata test set. The table shows the exact match depending on a hint of the target entity's appearance as a part of the model input.}
  \label{tab:target_inclusion}
  \begin{small}
  \begin{tabular}{lccr}
    \toprule
    Target position &  T5 & T5 + Neighbors & frequency\\
    \midrule
    In task         & 0.93 & 0.94 \up & 5 \%  \\
    In neighborhood & 0.34 & 0.82 \up & 8 \% \\
    Not in input    & 0.18 & 0.24 \up & 87 \% \\
    Total           & 0.23 & 0.32 \up & 100 \% \\
  \bottomrule
\end{tabular}
\end{small}
\vskip -0.1in
\end{table}

\section{Conclusions and Future work}
Our study found that incorporating KG neighborhood information into a generative model's context can enhance the quality of the link prediction task. We trained the T5-small model on transductive Wikidata5M and inductive ILPC datasets with various configurations and showed that coupling the node and its neighborhood information together better represents the node. Our approach significantly outperformed baselines on the largest real-world KGCs, with a comparable or fewer number of parameters. We also demonstrated that OpenAI's \texttt{gpt-3.5-turbo} performance on the KGC task also benefits from neighborhood information without prior fine-tuning. In addition to the quality improvement, we analyzed the impact of neighborhood imputation, highlighting the significance of selecting relevant triplets or extending the model's context in future works. Furthermore, future studies could explore the impact of model size, the effect of neighborhood imputation on the datasets from other domains, and neighborhood selection strategies.

\section*{Limitations}
Limitations of this study include the training of only relatively small models, with only T5 being employed. Further study of the benefits of larger models applied to KGC tasks should be performed. Additionally, each knowledge graph in this study was limited to only one model, which may not accurately represent more complex graphs that involve multiple datasets. Another limitation is that language models are better suited for real-world textual data and may not work well for KGC representing more specialized concepts, such as biology or medicine. Also, although neighborhood selection was performed, we cannot guarantee that the selected neighborhoods are optimal. Further works should include the way of extending the ability of models to capture larger parts of a graph or selecting the most relevant neighbors. Moreover, constructing prompts for testing GPT performance may not be optimal, and the model's maximal performance cannot be guaranteed without knowledge of its training data.

\section*{Ethics Statement}
As a field of study in the intersection of natural language processing and knowledge graphs, this work inherits the main ethical considerations of both of these domains. Automated graph completion tasks may produce various types of mistakes, including but not limited to factual and grammatical errors. Consequently, any application based on this work should employ robust error detection mechanisms prior to deployment. Additionally, given that our system employs a generative language model, there exists a risk of generating false or misleading content such as hallucinations and biases existing in pre-training and knowledge graph data. Users of the proposed method or any application based on it should be warned about the potential risks of using generative language models. Potential misuses may include purposeful disinformation and misleading language or factual information. Taking into account the aforementioned considerations, we believe that under ethically responsible usage this work will have a positive impact on artificial intelligence research and applications.

\section*{Acknowledgements}

We are grateful to Mikhail Galkin for valuable advice on knowledge graphs and introduction to the ILPC dataset. We are thankful to SberDevices for granting us access to additional computational resources.
A.C., A.B., and Y.K.'s work was supported by a grant for research centers in the field of artificial intelligence, provided by the Analytical Center for the Government of the Russian Federation in accordance with the subsidy agreement (agreement identifier 000000D730321P5Q0002) and the agreement with the Moscow Institute of Physics and Technology dated November 1, 2021 No. 70-2021-00138.

\bibliography{anthology,custom}
\bibliographystyle{acl_natbib}

\newpage
\appendix

\section{Input data pipeline}

\begin{figure}[h]
  \centering
  \includegraphics[width=\linewidth]{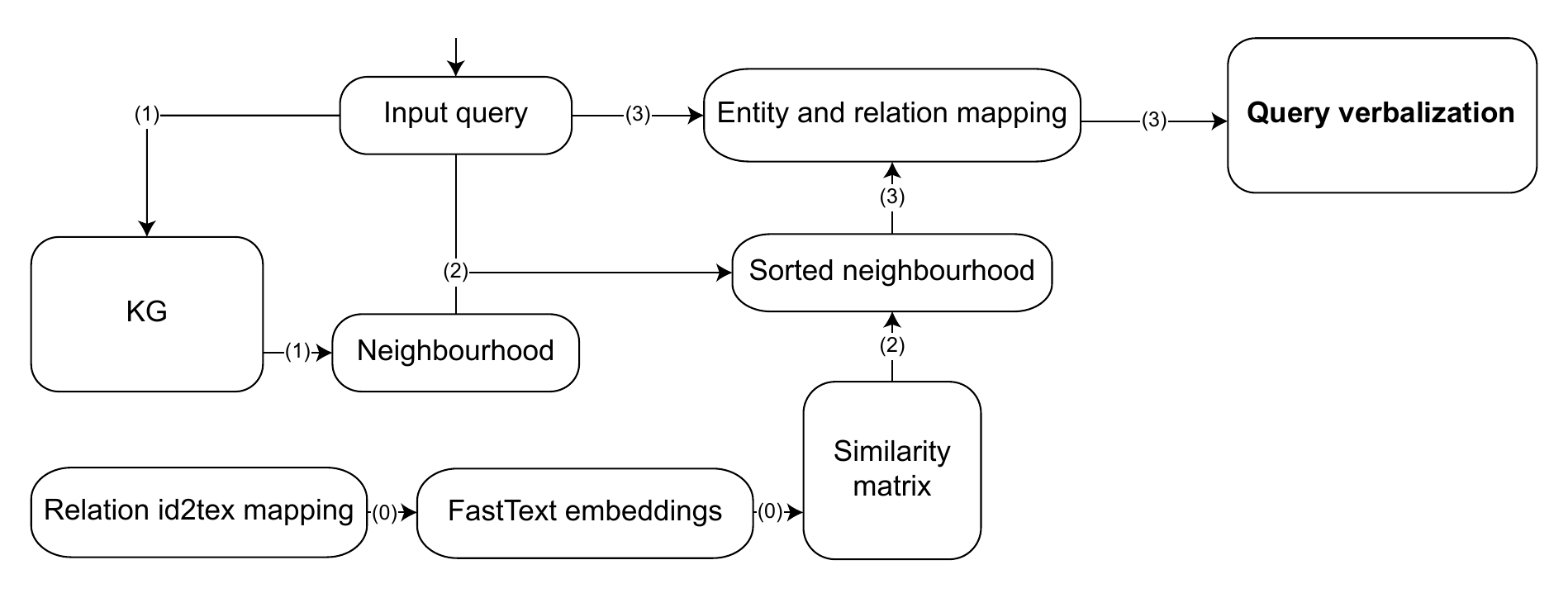}
  \caption{Visualization of the complete pipeline for the preparation of the input data of the language model: 0) pre-calculate FastText embeddings for each relation in KG and form a static similarity matrix out of them; 1) form a neighborhood for the input triplet; 2) sort the neighborhood by relation similarity to the relation in the input query; 3) verbalize and concatenate query triplet and its neighborhood.}
  \label{fig:full_pipeline}
\end{figure}

\section{Neighborhood formation}
\label{sec:neighborhood_formation}
\begin{figure}[h]

  \centering
  \includegraphics[width=\linewidth]{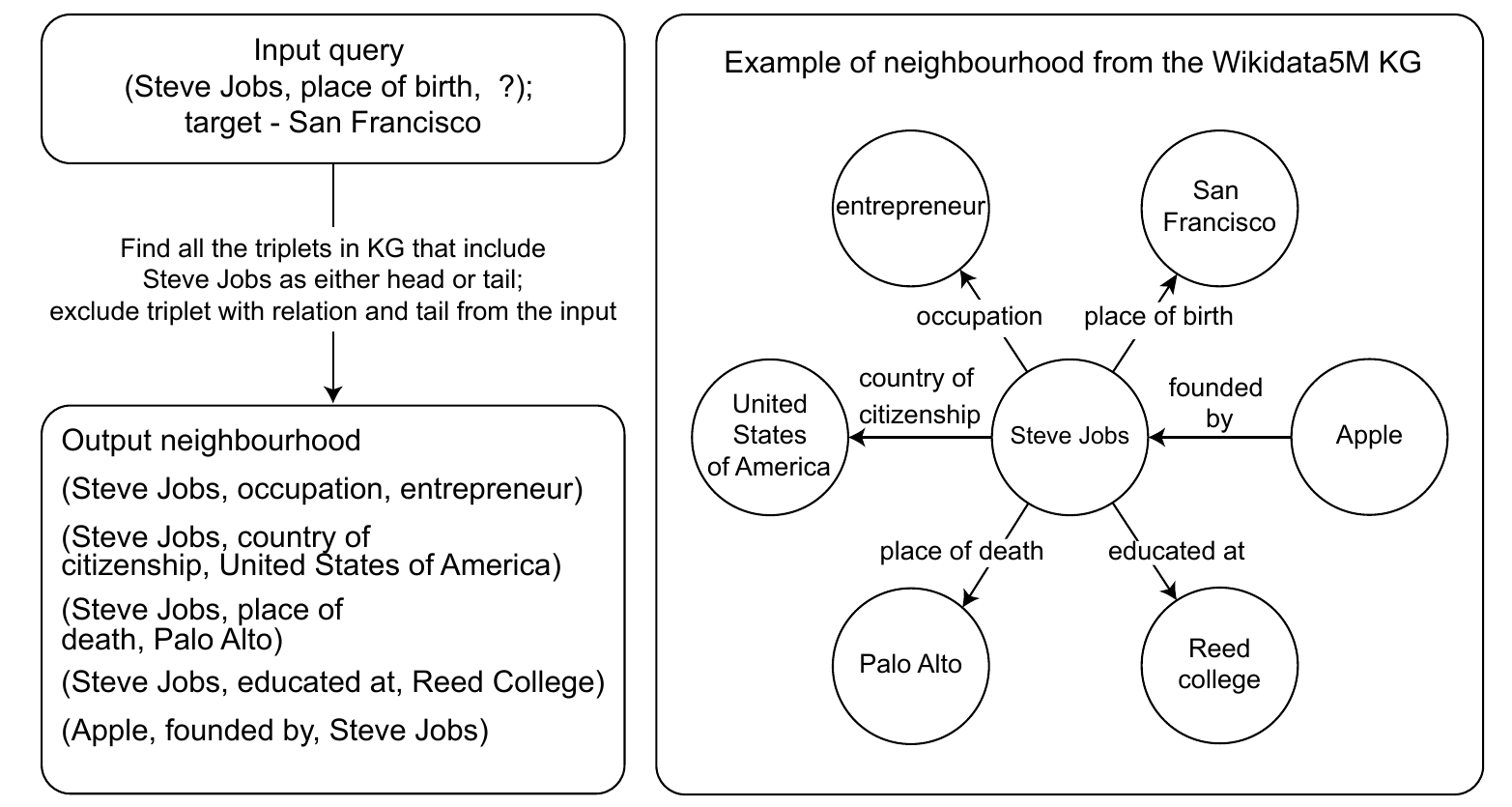}
  \caption{\textbf{Neighborhood formation} - taking the input triplet, search for the triplets in which the head from the input triplet is either head or tail, then exclude the triplet with relation and tail from the input. As a result, there is a list of triplets that form the neighborhood for a given triplet. The represented neighborhood from the Wikidata5m KG is incomplete and textually verbalized for ease of understanding.}
\label{fig:neighbourhood}
\end{figure}

To augment the semantic information contained in the entity and relationship names with the structural one, we used triplet neighborhoods. A triplet neighborhood is formed as follows: for a query ($head$, $relation$, ?) search for 1-hop neighbors associated with $head$ entity, while excluding from the formed neighborhood target $tail$ entity, which is associated with $head$ by $relation$; $head$ can be both head or tail entity in triplets that form this 1-hop neighborhood. In this manner, we obtain the context as a neighborhood of the graph, avoiding the leakage of the target entity. The process of neighborhood formation is illustrated on Figure \ref{fig:neighbourhood}.\\

\begin{figure}[h]
  \centering
  \includegraphics[width=\linewidth]{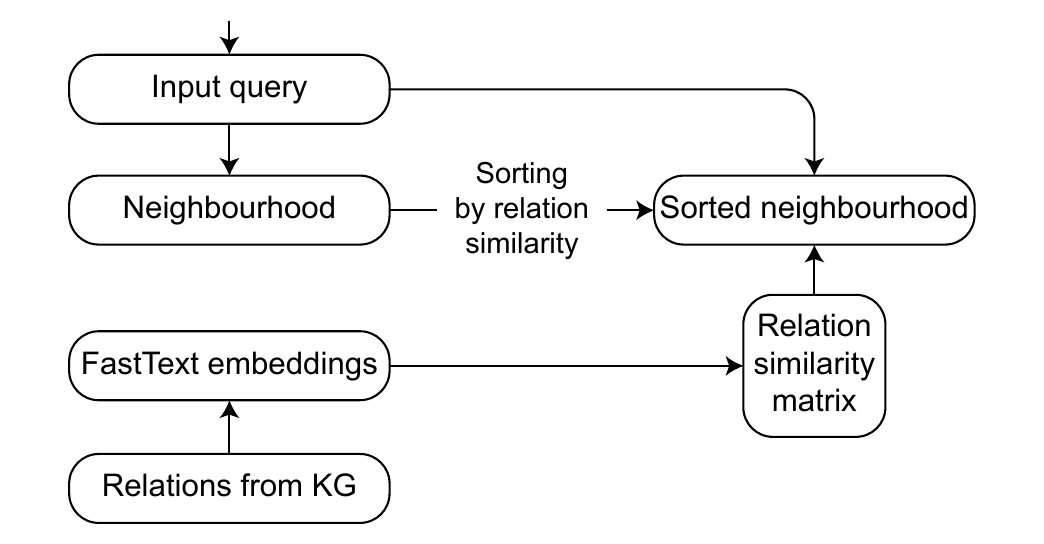}
  \caption{\textbf{Neighborhood sorting} - taking the input triplet, its neighborhood, and pre-calculated similarity matrix for verbalized relation embedded by the FastText model, sort the neighborhood by the similarity of relations in its triplets to the relation in the input query.}
  \label{fig:neighbourhood_sorting}
\end{figure}

However, some of the triplets from KGs result in extremely large neighborhoods, e.g. hub nodes such as "human". Therefore, we needed to tackle the problem of large neighborhoods to fit it into the model's context, which is limited to 512 tokens for most language models including T5. We propose to do it as follows: 1) sort the neighborhood by semantic similarity of relations from its triplets to the relation from the input triplet in order to prioritize more important information in the context; 2) limit the sorted neighborhood to 512 triplets, since this will always be at least as big as the size of the allowed context, and, after verbalization, specify the maximum length of 512 for the model tokenizer to fit the resulting verbalized neighborhood representation into the language model context. Neighborhood sorting by semantic proximity was performed using a pre-calculated matrix of cosine similarity of relations in KG, for similarity calculation the relations were embedded by the fasttext model~\citep{fasttext}. Figure \ref{fig:neighbourhood_sorting} displays a pipeline for neighborhood sorting. \\

Neighborhood composition was a relatively straightforward task for a smaller graph like ILPC, but for Wikidata KG with a large number of triplets in the graph, neighborhood formation took a significant amount of time and RAM. To solve this problem, we used the MongoDB database\footnote{https://www.mongodb.com}, which stored knowledge graph triplets in its collection in JSON format and contained indices for 'head' and 'tail' fields to speed up retrieving. In order not to store verbalized data in memory either during the verbalization process or during model training, verbalized representations were outputted to the MongoDB collection too, and retrieved by batches from the database during training. \\
\\

\section{Triplets verbalization}
\label{sec:triplets_verbalization}
\begin{figure}[h]
  \centering
  \includegraphics[width=\linewidth]{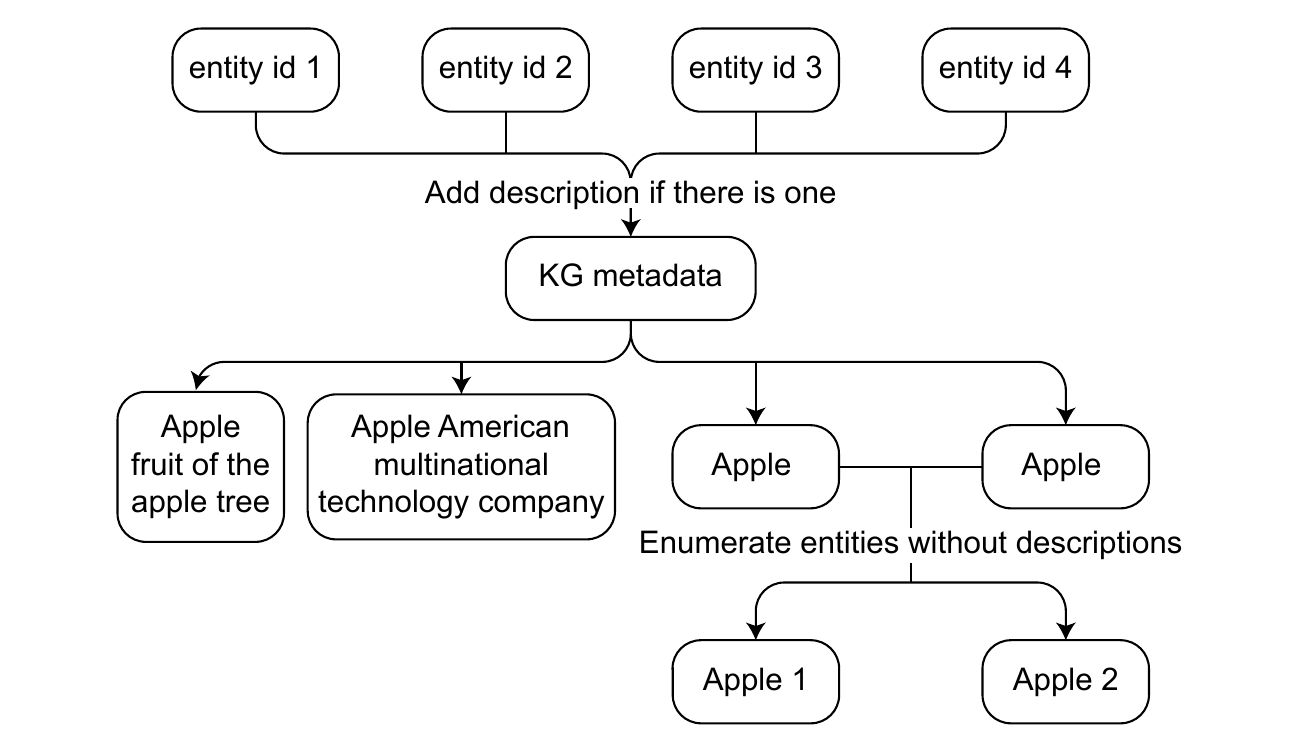}
  \caption{Disambiguation schema for wikidata5m entities: 1) concatenate entities' textual representation with corresponding description if there is one in the KG metadata; 2) if there is no description for some entities - concatenate their names with unique numerical id.}
  \label{fig:disambiguation}
\end{figure}

To pass the resulting representations of queries to the input of the language model, we had to perform a transformation of these queries and their neighborhoods into textual form. In order to do this, we first needed to generate a mapping of entity and relationship Wikidata IDs into corresponding texts. For all the datasets we performed our experiments on there is such textual mapping. However, often there is no unambiguous encoding in such mappings because many entities have the same name. The problem of non-unique textual representations concerns Wikidata KG entities, not relations, e.g. apple as a fruit of the apple tree and Apple as an American multinational technology company. To tackle this issue we used the same method KGT5 authors proposed: 1) concatenate non-unique entity names with corresponding descriptions if ones exist; 2) in case there is no description or after description, concatenation text representations cannot be disambiguated, add unique numeric ids. For all other smaller datasets we only added a unique numeric identifier at the end of the name of non-unique entities. The full disambiguation pipeline is displayed in Figure \ref{fig:disambiguation}. No such transformations were required for KG relations, as they were fully distinguishable by their text representation. \\
\label{sec:appendix}
\\ 
The next step after obtaining 1to1 mapping of entities and relation IDs to their text representations was verbalization of the resulting queries and corresponding neighborhoods in the KG. Textual representations of the query $(head_q, relation_q, ?)$ from KG were formed as follows: 
\begin{enumerate}
\item first part of query verbalization was formed via concatenation of $head_q$'s and $relation_q$'s text mappings with "predict" string, e.g. "predict Steve Jobs place of birth";
\item for each i-th triplet $(head_q, relation_i, tail_i)$ or $(head_i, relation_i, head_q)$ from the sorted query's neighborhood we formed a string consisting of $relation_i$ and entity not belonging to the input query, i.e. for query $(Steve\ Jobs, place\ of\ birth, ?)$ the verbalized triplet $(Steve\ Jobs, occupation, entrepreneur)$ from its neighborhood will be "occupation entrepreneur"; in case of so-called inverse triplet, e.g. in triplet $(head_i, relation_i, head_q)$, where $head_q$ is a tail of the neighborhood triplet, we add "inverse of " to the textual mapping of $relation_i$, i.e. for $(Apple, founded\ by, Steve\ Jobs)$ verbalization will be "inverse of founded by Apple";
\item at the end we concatenate the verbalization obtained in stage (1) and all the verbalizations from stage (2) using the" [SEP] " string inserted between all the above-mentioned strings. 
\end{enumerate}

\begin{figure}[h]
  \centering
  \includegraphics[width=\linewidth]{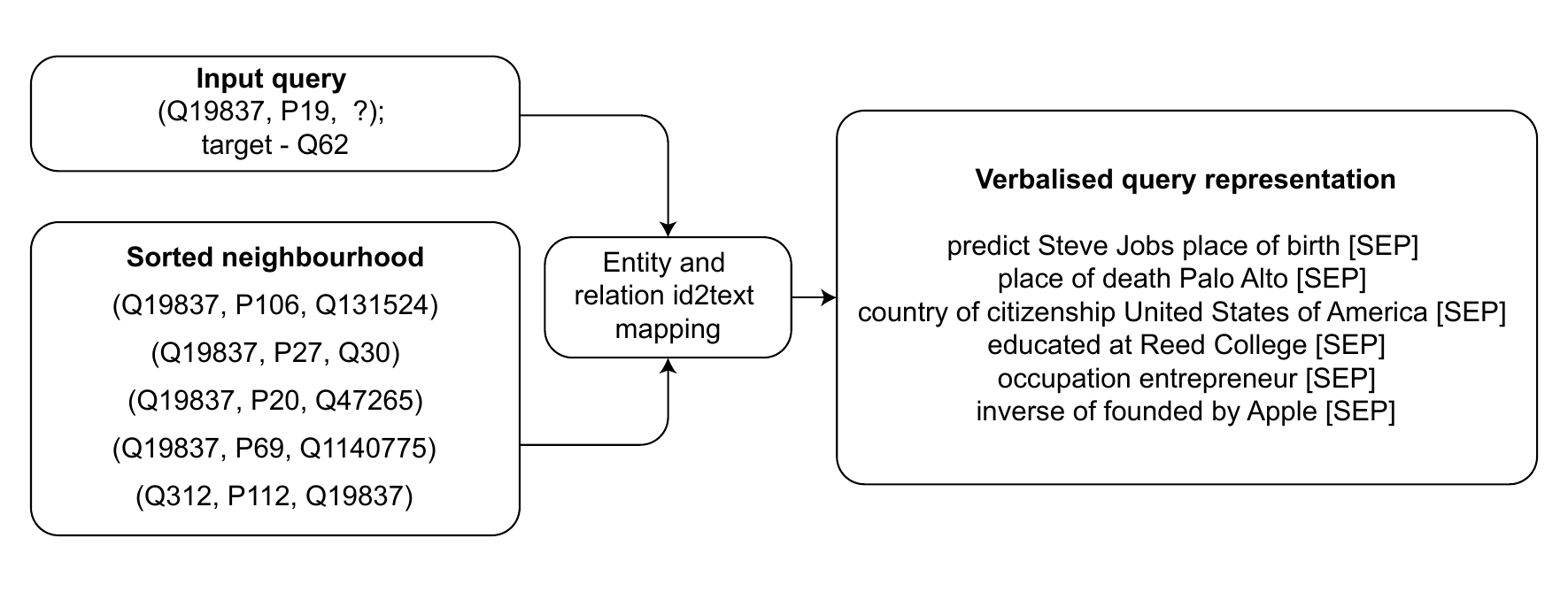}
  \caption{\textbf{Query verbalization} - map each entity and relation id from query triplet and its neighborhood triplets with corresponding textual representations; concatenate "predict" string, obtained query verbalization excluding target and the sorted triplets' verbalization without head entity from the query, separating each verbalized triplet by "[SEP]" token. }
  \label{fig:verbalization}
\end{figure}

Example of query and its neighborhood verbalization is represented on a Figure \ref{fig:verbalization}.

\section{Link prediction inference}
\label{sec:inference}

As in KGT5~\citep{kgt5}, to evaluate top-k possible answers, we sample k most probable sequences from the decoder and assign them their corresponding probabilities, which is a sum of log probabilities of each token in the sequence. Then, we assign $-\infty$ score for all the nodes that were not generated by the sampling procedure. In this way, we are producing results comparable with traditional KG models as we obtain plausibility scores for all possible triplets in KG. By such procedure, we can obtain whether a real score of the model (if the target sequence persists in k generated sequences) or the score that will be worse than a real (if the target sequence was not generated as one in the top-k probable sequences), thus not overestimating the resulting metric even though we do not score all the possible triplets. Additionally, to exclude both entities that do not belong to KG and known links from the train and valid graph in a transductive setting, we performed a filtering procedure specific to KGC evaluation protocol as described in the previous studies \citep{transe}. 

As a main metric for choosing the best model during evaluation, we used Exact Match (EM) or in other words Hits@1, which returns 1 in case the generated and target sequences are exactly the same or 0 otherwise. However, an exact match between the target and the generated sequence can be a rather strict metric for long target entity names, and for previously unseen entities. Thus, we came up with a new approach for scoring entities produced by generative models such as T5 for inductive KGC. In this approach, we first embedded verbalized entities from the inference part of KG by Sbert model~\citep{sbert} and then indexed the resulted vectors using FAISS index~\citep{faiss}. In such a way, we used a simple flat FAISS index to perform a lookup for the k most similar entities from inference KG to the generated one. Using the index for the inductive ILPC dataset, we retrieve entities from the KG that are semantically closest to the generated sequence even if the generated string is imperfect, but contains the right tokens referring to the target entity. In this manner, we could calculate Hits@k retrieving top-k closest to the generated entity names in a similar way as distance-based methods. Overall, in the case of inductive setup, we used a combination of exact match (EM) and FAISS-based metrics Hits@k calculated by the following formula: $1000EM + Hits@1$. In this way, we give more weight to the exact match metric, but in the case of equal EM in different checkpoints, the model with the best Hits@1 will be chosen. 

\section{Datasets statistics}
\label{sec:dataset_stat}

\begin{table}[h]
    
  \caption{Statistics of used KGs.}
  \label{tab:freq}
  \begin{center}
  \begin{small}
  \begin{tabular}{lcccl}
    \toprule
    Dataset &  \# Entities & \# Relations & \# Triplets \\
    \midrule
    Wikidata5M & 4.8M & 828 & 21M \\
    ILPC-small & 10,230 & 96 & 78,616 & \\
    ILPC-large & 46,626 & 130 & 202,446 \\
  \bottomrule
\end{tabular}
\end{small}
\end{center}
\label{table:dataset_statistics}
\end{table}

\section{Attention map interpretation}
\label{sec:attention_maps}

\begin{figure}[h]
  \centering
  \includegraphics[width=\linewidth]{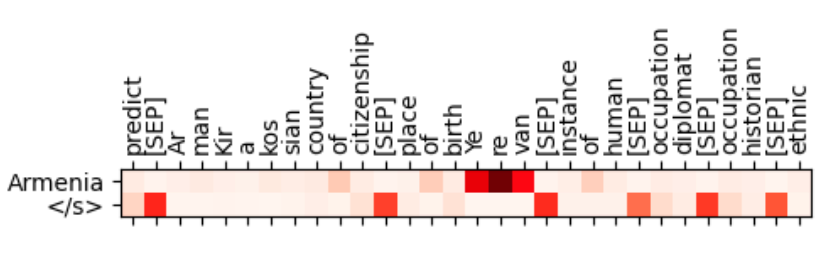}
  \caption{\textbf{Example of cross-encoder attention map} highlighting the importance of neighborhood triplets from the input}
  \label{fig:attention_maps}
\end{figure}

Figure \ref{fig:attention_maps} demonstrates a cross-encoder attention map highlighting the input neighborhood's significance for the target prediction. The query \textit{(Arman Kirakossian, country of citizenship, ?)} identifies Yerevan, Armenia's capital, as the birthplace of the target entity, thereby amplifying its role in predicting the right country of citizenship.

\section{Zero-shot link prediction with \texttt{gpt-3.5-turbo}}
\label{sec:gpt}
Since OpenAI API only returns generated text, we cannot get its probabilities. Therefore, we use a slightly different evaluation pipeline from other models. We generate up to three predictions with \texttt{gpt-3.5-turbo-0613}. If any of the randomly selected $k$ predictions match the target entity after cleaning and normalization, we set that model prediction as correct (Hits@$k$ in Table~\ref{tab:chatgpt}).

We use the following prompts to control the \texttt{gpt-3.5-turbo-0613}.

\subsection{Zero-shot link prediction without neighborhood}
System prompt:
\textit{
You will be provided with a incomplete triplet from the Wikidata knowledge graph. Your task is to complete the triplet with a tail (subject) based on the given triplet head (object) and relation. Your answer must only include the tail of the triplet with prefix 'Tail:'. Do not include relation into the answer.}

User prompt: \textit{Triplet to complete: <verbalized triplet>.}

\subsection{Zero-shot link prediction with neighborhood}
System prompt:
\textit{
You will be provided with a incomplete triplet from the Wikidata knowledge graph. Your task is to complete the triplet with a tail (subject) based on the given triplet head (object) and relation. Triplet to complete IS NOT in the list of related nodes, but it may contain helpful clues. Your answer must only include the tail of the triplet with prefix 'Tail:'. Do not include relation into the answer.}

User prompt: \textit{Adjacent relations: <verbalized neighborhood>\textbackslash nTriplet to complete: <verbalized triplet>.}

\section{Overview of concurrent work}
While our work was under review, the concurrent work KGT5-context~\cite{kochsiek2023friendly} was published. Following the same idea, KGT5-context uses verbalized node neighborhood as additional context for KGC with the KGT5 model. However, KGT5-context explores the applicability of neighborhood information for KGC only in transductive setup. Also, neighborhood sampling in KGT5-context was performed randomly instead of prioritizing the neighbors, which makes the results less reproducible and does not consider the importance of neighbors in the context of a query. Furthermore, we show that this approach generalizes well to other pre-trained language models like \texttt{gpt-3.5-turbo} in a zero-shot setting without pre-training on KGC datasets.

We have made the effort and tried our best to reproduce the results of the KGT5 model~\cite{kgt5} on Wikidata5M. We report the results of \textit{T5-small not pre-trained} (Table~\ref{tab:results}), which is our version of the KGT5 model. Therefore, all our other models were trained and evaluated in the same setting and with the same pipeline as for \textit{T5-small not pre-trained} model.

\end{document}